\pdfoutput=1
\documentclass[11pt,a4paper]{article}
\usepackage[final]{naacl2021}
\usepackage{times}
\usepackage{latexsym}

\usepackage{tabularx}
\usepackage{graphicx}
\usepackage{float}
\usepackage{makecell}
\usepackage{hyperref}
\usepackage{xurl}

\usepackage{microtype}

\usepackage{times}
\usepackage{latexsym}

\usepackage{float}

\usepackage[utf8]{inputenc} 
\usepackage[T1]{fontenc}    
\usepackage{hyperref}       
\usepackage{url}            
\usepackage{booktabs}       
\usepackage{amsfonts}       
\usepackage{nicefrac}       
\usepackage{microtype}      
\usepackage{graphicx}
\usepackage{float}
\usepackage{amssymb}
\usepackage{amsmath}
\usepackage{mathtools}
\usepackage{bbm}
\usepackage{amsthm}
\usepackage{xcolor}
\usepackage{pifont}
\usepackage{subcaption}
\usepackage{tabularx}

\newtheorem*{definition*}{Definition}

\renewcommand{\mid}{{\kern 0.075em}|{\kern 0.075em}}
\DeclareMathOperator*{\argmax}{arg\,max}

\usepackage{multirow}

\usepackage{amsmath,amssymb}

\begin{document}

\title{\textit{Nutribullets Hybrid}: Multi-document Health Summarization}

 \author{Darsh J Shah\textsuperscript{1} ~~
Lili Yu\textsuperscript{2}~~
Tao Lei\textsuperscript{2} ~~
Regina Barzilay\textsuperscript{1}\\
\textsuperscript{1}{Computer Science and Artificial Intelligence Lab, MIT}\\
\textsuperscript{2}{ASAPP, Inc.}\\
\small{darsh@csail.mit.edu ~~
liliyu@asapp.com ~~
tao@asapp.com ~~
regina@csail.mit.edu}}
 
\date{}

\maketitle
\begin{abstract}
We present a method for generating comparative summaries that highlights similarities and contradictions in input documents. The key challenge in creating such summaries is the lack of large parallel training data required for training typical summarization systems. To this end, we introduce a hybrid generation approach inspired by traditional concept-to-text systems. To enable accurate comparison between different sources, the model first learns to extract pertinent relations from input documents. The content planning component uses deterministic operators to aggregate these relations after identifying a subset for inclusion into a summary. The surface realization component lexicalizes this information using a text-infilling language model. 
By separately modeling content selection and realization, we can effectively train them with limited annotations. We implemented and tested the model in the domain of nutrition and health -- rife with inconsistencies. Compared to conventional methods, our framework leads to more faithful, relevant and aggregation-sensitive summarization -- while being equally fluent.\footnote{Our code and data is available at \url{https://github.com/darsh10/Nutribullets}}
\end{abstract}

\section{Introduction}

\begin{figure}[!t!]
\includegraphics[width=1.0\columnwidth]{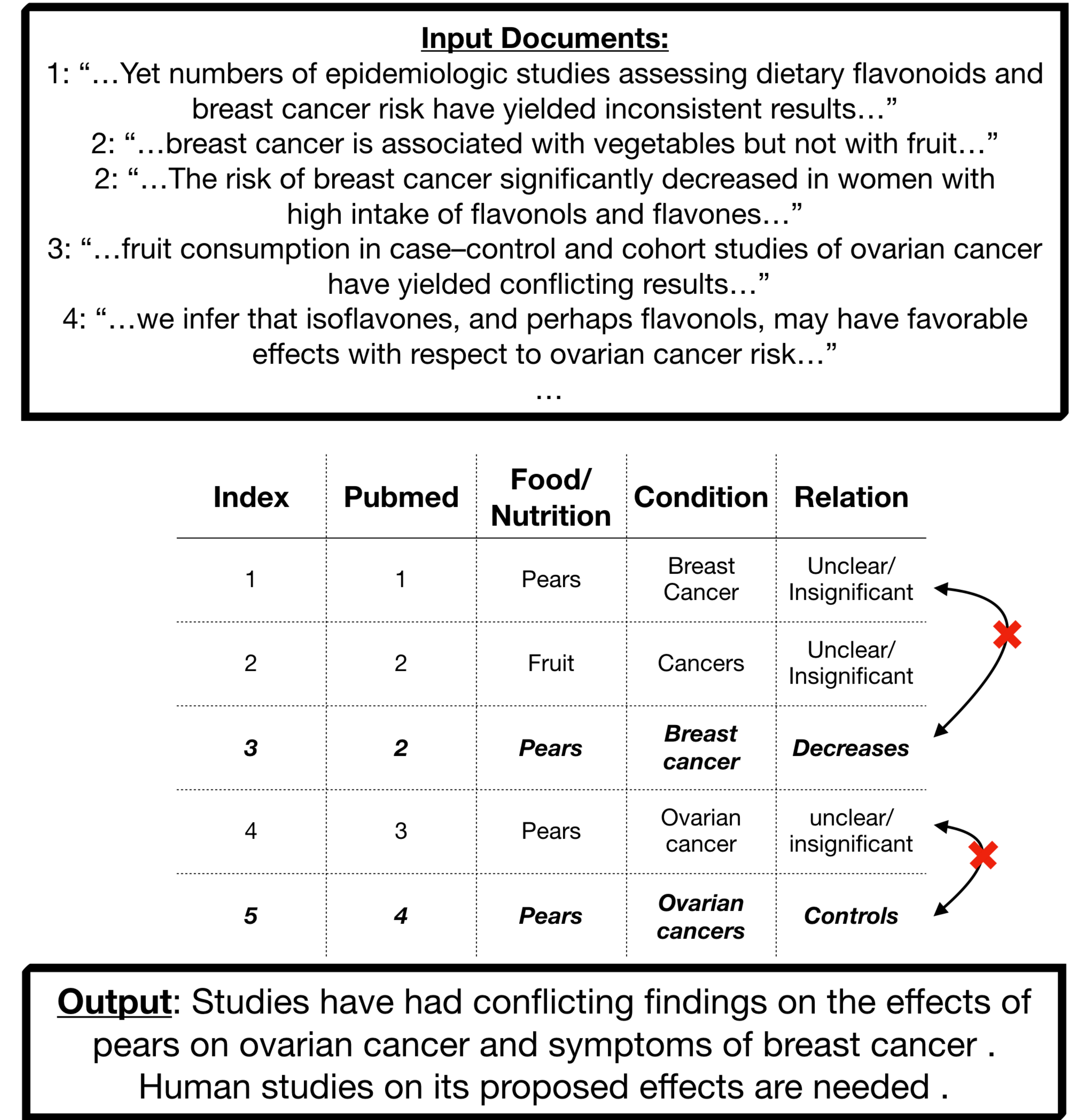}
\caption{\small{We consider the database extracted from four Pubmed studies on Pears and Cancer. The key facts (\emph{bold}) and consensus (\emph{contradiction}) are realized in the text generated by our model.}
\label{fig:example-1}}
\end{figure}

Articles written about the same topic rarely exhibit full agreement. To present an unbiased overview of such material, a summary has to identify points of consensus and highlight contradictions. For instance, in the healthcare domain, where 
studies often exhibit wide divergence of findings, such comparative summaries are generated by human experts for the benefit of the general public.\footnote{Examples include \url{https://www.healthline.com} and \url{https://foodforbreastcancer.com}.} Ideally, this capacity will be automated given a large number of relevant articles and continuous influx of new ones that require a summary update to keep it current. However, standard summarization architectures cannot be utilized for this task since the amount of comparative summaries is not sufficient for their training. 

In this paper, we propose a novel approach to multi-document summarization based on a neural interpretation of traditional 
concept-to-text generation systems. Specifically, our work is inspired by the symbolic multi-document summarization system of \citep{radev-mckeown-1998-generating} which produces summaries that explicitly highlight agreements, contradictions and other relations across input documents. While their system was based on human-crafted templates and thus limited to a narrow domain, our approach learns different components of the generation pipeline from data. 

To fully control generated content, we frame the task of comparative summarization as concept-to-text generation. As a pre-processing step, we extract pertinent entity pairs and relations (see Figure \ref{fig:example-1}) from input documents. The \emph{Content Selection} component identifies the key tuples to be presented in the final output and establishes their comparative relations (e.g., consensus) via aggregation operators. Finally, the \emph{surface realization} component utilizes  a text-infilling language model to translate these relations into a summary. Figure \ref{fig:example-1} exemplifies this pipeline, showing selected key pairs (marked in bold), their comparative relation -- \emph{Contradiction} (rows 1 \&3 and rows 4\&5 conflict), and the final summary.\footnote{We compare the selected content with other entries in the database, identifying two contradictions.}

This generation architecture supports refined control over the summary content, but at the same time does not require large amounts of parallel data for training. The latter is achieved by separately training content selection and content realization components. Since the content selection component operates over relational tuples, it can be robustly trained to identify salient relations utilizing limited parallel data. Aggregation operators are implemented using simple deterministic rules over the database where comparative relations between different rows are apparent. On the other hand, to achieve a fluent summary we have to train a language model on large amounts of data, but such data is readily available. 

In addition to training benefits,  this hybrid architecture enables human writers to explicitly guide content selection. This can be achieved by defining new aggregation operators and including new inference rules into the content selection component. Moreover, this architecture can flexibly support other summarization tasks, such as generation of updates when new information on the topic becomes available.

We apply our method for generating summaries of Pubmed publications on nutrition and health.  Typically, a single topic in this domain is covered by multiple studies which often vary in their findings making it particularly appropriate for our model. We perform extensive automatic and human evaluation to compare our method against state-of-the-art summarization and text generation techniques. While seq2seq models receive competent fluency scores, our method performs stronger on task-specific metrics including \emph{relevance}, \emph{content faithfulness} and \emph{aggregation cognisance}. Our method is able to produce summaries that receive  an absolute 20\%  more on aggregation cognisance, an absolute 7\% more on content relevance and 7\% on faithfulness to input documents than the next best baseline in traditional and update settings. 
\section{Related Work}
\textbf{Text-to-text Summarization}
Neural sequence-to-sequence models ~\citep{rush-etal-2015-neural, cheng-lapata-2016-neural,see-etal-2017-get} for document summarization have shown promise and have been adapted successfully for multi-document summarization~\citep{zhang-etal-2018-adapting, lebanoff2018adapting, baumel2018query, amplayo2019informative, multinews}. 
Despite producing fluent text, these techniques may generate false information which is not faithful to the original inputs~\citep{puduppully2019data,kryscinski2019evaluating}, 
 especially in low resource scenarios. In this work, we are interested in producing faithful and fluent text cognizant of aggregation amongst input documents, where few parallel examples are available.

Recent language modeling approaches \citep{devlin2018bert,stern2019insertion,shen2020blank,donahue2020enabling} can also be extended for text completion. Our work is a text-infilling language model where we generate words in place of relation specific blanks to produce a faithful summary.

Prior work \citep{mueller2017sequence,fan2017controllable,guu-etal-2018-generating} on text generation also control aspects of the produced text, such as style and length. While these typically utilize tokens to control the modification, using prototypes to generate text is also very common \citep{guu-etal-2017-language,li-2018-learning,shah2019automatic}. In this work, we utilize aggregation specific prototypes to guide aggregation cognizant surface realization.

\begin{figure*}[!t]
\centering
\includegraphics[width=\textwidth]{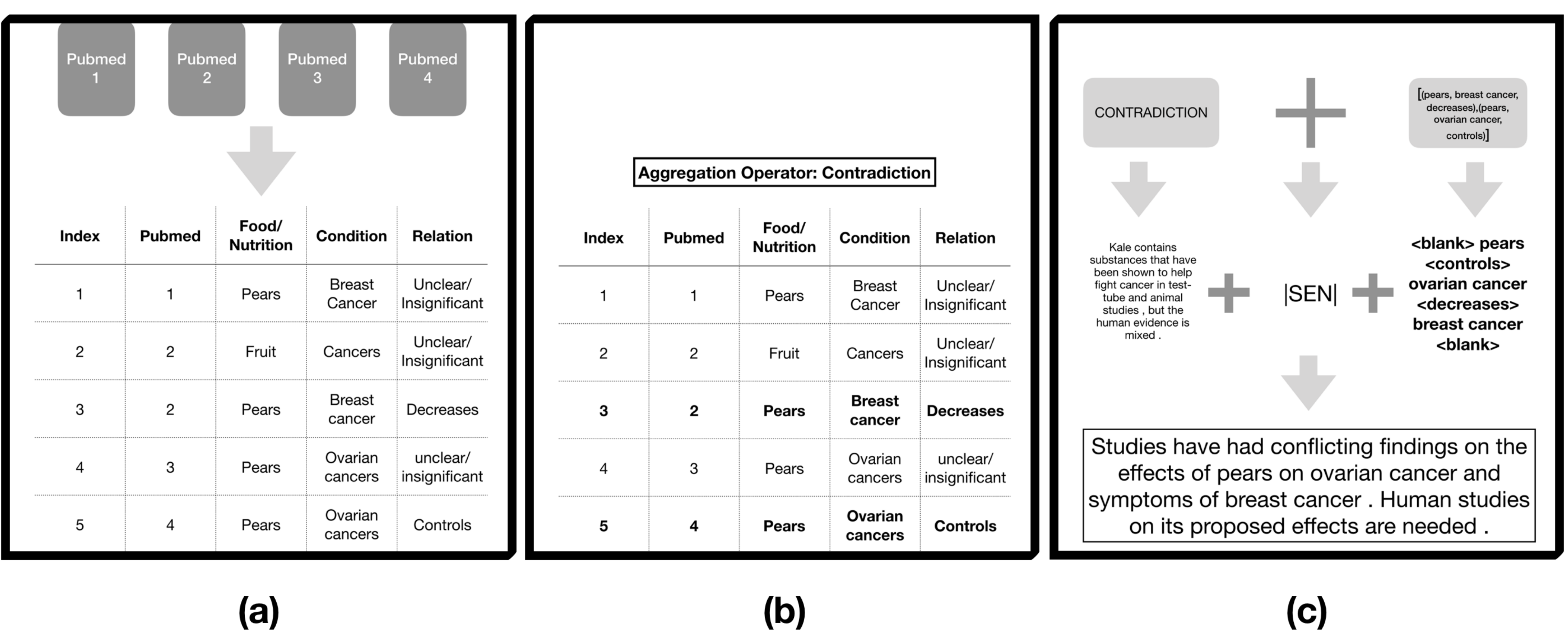}
\caption{Illustrating the flow of our \emph{Nutribullets Hybrid} system. In this example, our model takes in four Pubmed studies to produce a database (a). The \emph{Content Selection} model selects two tuples (bold) and identifies the aggregation operator as Contradiction (b). Finally, the \emph{Surface Realization} model takes in the tuples and aggregation operator to produces a summary which is faithful to input entities and aggregation cognizant (c).}
\label{fig:mask}
\end{figure*}

\textbf{Data-to-text Summrization} Traditional approaches for data-to-text generation have operated on symbolic data from databases. \citet{mckeown1995generating,radev-mckeown-1998-generating,barzilay-etal-1998-new} introduce two components of content selection and surface realization. Content selection identifies and aggregates key symbolic data from the database which can then be realized into text using templates. Unlike modern data-to-text systems \citep{wiseman-etal-2018-learning,  puduppully2019data,sharma-etal-2019-entity,wenbo2019concept} these approaches capture document consensus and aggregation cognisance. While the neural approaches alleviate the need for human intervention, they do need an abundance of parallel data, which are typically from one source only. Hence, modern techniques do not deal with input documents' consensus in low resource settings.

\section{Method}
\label{sec:Method}

\begin{table*}[t!]
\center
\scalebox{0.8}{
\begin{tabular}{c|c}
\toprule
Aggregation Operator & Deterministic Rule\\
\midrule
 Under-Reported &   $|$Pubmed Studies$|$ < Threshold\\
 Population Scoping &   $|$Specific Population$|$ < Threshold\\
 Contradiction &  $(e^m_1 == e^n_1 \&\& e^m_2 == e^n_2 \&\& r^m != r^n)$ for any two tuples $m,n$ from different studies \\
 Agreement & None of the Above \\
\bottomrule
\end{tabular}
}
\caption{Deterministic Rules to identify the Aggregation Operator.}
\label{table:rules}
\end{table*}

Our goal is to generate a text summary $y$ for a food from a pool of multiple scientific abstracts $X$. In this section, we describe the framework of our  \emph{Nutribullets Hybrid} system, illustrated in Figure \ref{fig:mask}.

\subsection{Overview}

 We attain food health entity-entity relations, for both input documents $X$ and the summary $y$, from entity extraction and relation classification modules trained on corresponding annotations (Table \ref{table:entity_relations}). \newline
\textbf{Notations:} For $N$ input documents, we collect $X_{\mathcal{G}} =  \{\mathcal{G}^x_p\}_{p=1}^{N}$, a database of entity-entity relations $\mathcal{G}^x_p$. $\mathcal{{G}}_p = (e_1^k,e_2^k,r^k)_{k=1}^{K}$ is a set of $K$ tuples of two entities $e_1$, $e_2$ and their relation $r$. $r$ represents relations such as the effect of a nutrition entity $e_1$ on a condition $e_2$ (see Table \ref{table:entity_relations}).\footnote{We train an entity tagger and relation classifier to predict $\mathcal{G}$ and also for computing knowledge based evaluation scores. More details on models and results are shared later.}We have raw text converted into symbolic data.

Similarly, we denote the corpus of summaries  as $Y = \{(y_m, \mathcal{G}^y_m, O^y_m)_{m=1}^{M}\}$, where $y_m$ is a concise summary, $\mathcal{G}^y_m$ is the set of entity-entity relation tuples and $O^y_m$ is the realized aggregation, in $M$ data points.\newline
\textbf{Modeling:} Joint learning of content selection, information aggregation and text generation for multi-document summarization can be challenging. This is further exacerbated in our technical domain with few parallel examples and varied consensus amongst input documents. To this end, we propose a solution using Content Selection and Aggregation and Surface Realization models.

Raw text from $N$ input documents is converted into a mini-database $X_{\mathcal{G}}$ of relation tuples.  The content selection and aggregation model operates on such symbolic data. We use $X_\mathcal{G}$ and $Y$ to train the content selection model. During inference, we identify from $X_{\mathcal{G}}$ a subset $C$ of content to present in the final output. In order to produce a summary cognizant of consensus amongst inputs, we identify the aggregation operator $O$ based on $C$ and other relevant tuples in $X_{\mathcal{G}}$.

The surface realization model produces a relevant, faithful and aggregation cognizant output. The model is trained only using $Y$. During inference, the model realizes text using the selected content $C$ and the aggregation operator $O$.

\subsection{Content Selection and Aggregation}
Our content selection model takes a mini-database of entity-entity relation tuples $X_{\mathcal{G}}$ as input, and outputs the key tuples $C$ and  the aggregation operator $O$.

Content selection and aggregation consists of two parts -- (i) identifying key content $P(C|X_{\mathcal{G}})$ and (ii) subsequently identifying the aggregation operator $O$ using  $C,X_{\mathcal{G}}$. \newline
\textbf{Content Selection}
Identifying key content involves selecting important, diverse and representative tuples from a database. While clustering and selecting from the database tuples is a possible solution, we model our content selection as a finite Markov decision process (MDP). This allows for an exploration of different tuple combinations while incorporating delayed feedback from various critical sources of supervision (similarity with target tuples, diversity amongst selected tuples etc). We consider a multi-objective reinforcement learning algorithm \citep{williams1992simple} to train the model. Our rewards (Eq. \ref{eq:multi-obj-rewards}) allow for the selection of informative and diverse relation tuples. 

The MDP's state is represented as $s_t= (t,\{c_1, \dots, c_t\}, \{z_1, z_2, ..., z_{m-t}\})$ where $t$ is the current step, $\{c_1, \dots, c_t\}$ is the content selected so far and $\{z_1, z_2, ..., z_{m-t}\}$ is the remaining entity-entity relation tuples in the $m$-sized database. The action space is all the remaining tuples plus one special token, $Z\cup \{\textit{STOP}\}$.\footnote{STOP and NEW LIST get special embeddings.} The number of actions is equal to $|m-t|+1$. As the number of actions is variable yet finite, we parameterize the policy $\pi_\theta (a | s_t)$ with a model $f$ which maps each action and state $(a,s_t)$ to a score, in turn allowing a probability distribution over all possible actions using softmax. At each step, the probability that the policy selects $z_i$ as a candidate is:
\begin{equation}
\pi_\theta (a\!=\!z_i|s_t) = \frac{\exp(f(t,\hat{z_i}, \hat{c_i*}))}{\sum_{j=1}^{m-t+1}\exp(f(t, \hat{z_j}, \hat{c_j*}))}
\end{equation}

where $c_i* = \argmax_{c_j}(cos(\hat{z_i},\hat{c_j}))$ is the selected content closest to $z_i$, $\hat{z_i}$ and $\hat{c_i*}$ are the encoded dense vectors, $cos(u,v)=\frac{u \cdot v}{||u||\cdot ||v||}$ is the cosine similarity of two vectors and $f$ is a feed-forward neural network with non-linear activation functions that outputs a scalar score for each action $a$.

The selection process starts with $Z$. Our module iteratively samples actions from $\pi_\theta(a|s_t)$ until selecting \textit{STOP}, ending with selected content $C$ and a corresponding reward. We can even allow for the selection of partitioned tuple sets by adding an extra action of "NEW LIST", which allows the model to include subsequent tuples in a new group.

We consider the following individual rewards:
\begin{itemize}
\setlength\itemsep{0.005em}
    \item $\mathcal{R}_e = \sum_{c \in C}cos(\hat{e_{1c}}, \hat{e_{1y}}) + cos(\hat{e_{2c}}, \hat{e_{2y}})$ is the cosine similarity of the structures of the selected content $C$ with the structures present in the summary $y$ (each summary structure accounted with only one $c$), encouraging the model to select relevant content.
    \item $\mathcal{R}_d = \mathbbm{1}[\max_{i,j}(cos(\hat{c_j},\hat{c_i})) < \delta] $ computes the similarity between pairs within  selected content $C$, encouraging the selection of diverse tuples.
    \item $r_p$ is a small penalty for each action step to encourage concise selection. 
\end{itemize}
The multi-objective reward is computed as
\begin{equation}
\label{eq:multi-obj-rewards}
\mathcal{R} = w_e\mathcal{R}_e \! +\!  w_d\mathcal{R}_d - |C|r_p , 
\end{equation}
where $w_e$, $w_d$ and $r_p$ are hyper-parameters.

During training the model is updated based on the rewards. During inference the model selects an ordered set of key and diverse relation tuples corresponding to appropriate health conditions.

\begin{table*}[t!]
\centering
\scalebox{0.68}{
\begin{tabular}{l|c|c|c|c}
\toprule
 Relation Type & \bf $e_1$      & \bf $e_2$    & \bf $r$& \bf Example   \\ 
\midrule
Causing  & \begin{tabular}[c]{@{}c@{}}Food, Nutrition \\ \end{tabular} & Condition   & \begin{tabular}[c]{@{}c@{}}Increase, Decrease, \\ Satisfy, Control, \\ Unclear/Insignificant\end{tabular}    & \begin{tabular}[c]{@{}c@{}}(tart cherry juice, melatonin levels, increase), \\ (water, daily fluid needs, satisfy)\end{tabular} 
\\ 
\midrule
Containing & Food, Nutrition  & Nutrition   & Contain          & (blueberries, antioxidants, contain)               \\
\bottomrule
\end{tabular}
}
\caption{Details of entity-entity relationships that we study and some examples of $(e_1, e_2, r)$}
\label{table:entity_relations}
\end{table*}

\textbf{Consensus Aggregation} Identifying the consensus amongst the input documents is critical in our multi-document summarization task. We model the aggregation operator of our \emph{Content Selection} using simple one line deterministic rules as shown in Table \ref{table:rules}. The rules are applied to the key $C$ entity-entity relation pairs in context of $X_\mathcal{G}$. In our example in Figure \ref{fig:example-1}, $O$ is Contradiction because of rows 1\&3 and rows 4\&5 (rows 1\&3 only would also make it Contradiction).

\subsection{Surface Realization}
The surface realization model $P(y|O,C)$, performs the critical task of generating a summary guided by both the entity-entity relation tuples $C$ and the aggregation operator $O$. The model allows for robust, diverse and faithful summarization compared to traditional template and modern seq2seq approaches. 

We propose to model this process as a prototype-driven text infilling task. The entities from $C$ are used as fixed tokens with relations as special blanks in between these entities. This is prefixed by a prototype summary corresponding to $O$. For the example shown in Figure \ref{fig:mask}, we concatenate using $|SEN|$ a randomly sampled contradictory summary \textit{"Kale contains substances ... help fight cancer ... but the human evidence is mixed ."} to $C$ \textit{"<blank> pears <controls> ovarian cancer <decreases> breast cancer <blank>"}. The infilling language model produces text corresponding to relations between entities while maintaining an overall structure which is cognizant of $O$. \footnote{Summaries in our training data are labelled with $O^y_m$ as belonging to one of the four categories of \emph{Under-reported, Population Scoping, Contradiction or Agreement} to accommodate such training.}

The model is trained on the few sample summaries from the training set using $\mathcal{G}^y_m$ and $O^y_m$ to produce $y_m$. Providing aggregation and content guidance during generation alleviates the low-resource issue.
\section{Summary and Update Setting}
\label{sec:settings}
In this section we describe the setting of summary updates. In a real world setting, we would often receive new input documents such as scientific studies about the same subject which necessitate a change in an old summary.

In context of our food and health summarization task, the goal is to update an old summary about a food and health condition on receiving results from new scientific studies from Pubmed. Our model can accommodate this scenario fairly easily. We describe the minor changes to the \emph{Content Selection and Aggregation} and \emph{Surface Realization} models for such a setting. 

We are provided an original summary and can extract it's content $C'$ and can also construct the mini-database $X_{\mathcal{G}}$ from the text of the new documents. We identify the aggregation between the new studies' $X_{\mathcal{G}}$ and original summary's content $C'$ first. Depending on the aggregation identified, corresponding content $C$ is selected from $X_{\mathcal{G}}$. For instance, in case of a contradiction, we are keen on identifying content leading to this contradiction. The subsequent \emph{Surface Realization} is dependent on $O$, the selected $C$ and the $C'$ present in the original summary ($P(y|O,C+C')$).

\section{Experiments}
\begin{table*}[!htbp]
\small
\centering
\scalebox{0.9}{
\begin{tabular} {l|c|c|c|c|c|c} 
\toprule
\multicolumn{1}{c}{}&\multicolumn{4}{c}{Automatic Evaluation} & \multicolumn{2}{c}{Human Scores}\\
\textsc{Model}              & \textsc{RougeL} & \textsc{KG(G)} & \textsc{KG(I)} & \textsc{Ag} &  \textsc{Relevance} & \textsc{Fluency} \\ 
\midrule
Copy-gen             & 0.12 & 0.21 & 0.50 & 0.64 & 1.93 & 1.89 \\
GraphWriter         & 0.14 & 0.03 & 0.69 & 0.64 & 1.86 & 2.76\\
Entity Data2text & 0.16 & 0.13 & 0.57 & 0.67 & 2.03 & 3.43\\
Transformer          & \textbf{0.20}  & 0.21 & 0.64 & 0.67 & 2.66 & \textbf{3.76} \\
\midrule
Ours        & 0.18  & \textbf{0.30} & \textbf{0.76} & \textbf{0.89} & \textbf{3.03} & 3.46 \\
\bottomrule
\end{tabular}
}
\caption{Automatic evaluation -- Rouge-L score (RougeL), KG in gold(G), KG in input(I) and Aggregation Cognisance (Ag) in our model and various baselines in the single issue setting, is reported. Human evaluation on Relevance and Fluency, on 1-4 Likert scale from 3 annotators, is also reported. The best results are in \textbf{bold}.}
\label{tab:automatic-full}
\end{table*}
\textbf{Dataset} 
We utilize a real world dataset for Food and Health summaries, crawled from \url{https://www.healthline.com/nutrition} \citep{shah2021nutri}. The HealthLine dataset consists of scientific abstracts as inputs and human written summaries as outputs. The dataset consists of 6640 scientific abstracts from Pubmed, each averaging 327 words. The studies in these abstracts are cited by domain experts when writing summaries in the Healthline dataset, forming natural pairings of parallel data. Individual summaries average 24.5 words and are created using an average of 3 Pubmed abstracts. Each food has multiple bullet summaries, where each bullet typically talks about a different health impact (hydration, diabetes etc). We assign each food article randomly into one of the train, development or test splits. Entity tagging and relation classification annotations are provided for the Pubmed abstracts and the healthline summaries. 
\newline
\textbf{Settings:} 
We consider three settings. \newline
    \textbf{1. Single Issue:} We use the individual food and health issue summaries as a unique instance of food and single issue setting. We split 1894 instances  80\%,10\%,10\% to train, dev and test. \newline
    \textbf{2. Multiple Issues:} We group each food's article Pubmed abstract inputs and multiple summary outputs as a single parallel instance. 464  instances are split 80\%,10\%,10\% to train, dev and test. \newline
    \textbf{3. Summary Update:} We consider two kinds of updates -- new information is fused to an existing summary and  new information contradicts an existing summary. For fusion we consider single issue summaries that have multiple conditions from different Pubmed studies (bananas + low blood pressure from one study and bananas + heart health from another study). We partition the Pubmed studies to stimulate an update. The contradictory update setting is where we artificially introduce conflicting results in the input document set so that the aggregation changes from Agreement to Contradictory. We have a total of 103 test instances. All models are trained atop of Single issue data. 
    
\textbf{Evaluation} 
We evaluate our systems using the following automatic metrics. 
\textit{Rouge} is an automatic metric used to compare the model output with the gold reference \citep{lin2004rouge}.
\textit{KG(G)} computes the number of entity-entity pairs with a relation in the gold reference, that are generated in the output.\footnote{We run entity tagging plus relation classification on top of the model output and gold summaries. We match the gold $(e_i^{g},e_j^{g},r^{g})$ tuples using word embedding based cosine similarity with the corresponding entities in the output structures $(e_i^{o},e_j^{o},r^{o})$. A cosine score exceeds a threshold of 0.7 is set (minimize false positives) to identify a match.} This captures relevance in context of the reference.
\textit{KG(I)}, similarly,  computes the number of entity-entity pairs in the output that are present in the input scientific abstracts. This measures faithfulness with respect to the input documents.
\textit{Aggregation Cognisance (Ag)} measures the accuracy of the model in producing outputs which are cognizant of the right aggregation from the input, (Under-reported, Contradiction or Agreement). We use a rule-based classifier to identify the aggregation implied by the model output and compare it to the actual aggregation operator based on the input Pubmed studies.

In addition to automatic evaluation, we have human annotators score our models on  relevance and  fluency. Given a reference summary, \textit{relevance} indicates if the generated text shares similar information. \textit{Fluency} represents if the generated text is grammatically correct and written in well-formed English. Annotators rate relevance and fluency on a 1-4 likert scale ~\cite{albaum1997likert}. We have 3 annotators score every data point and report the average across the scores. \newline
\textbf{Baselines}
In order to demonstrate the effectiveness of our method, we compare it against text2text and data2text state-of-the-art (\emph{sota}) methods.\newline
     \textbf{Copy-gen (Text2text):} \citet{see-etal-2017-get} is a \emph{sota} technique for summarization, which can copy from the input or generate words.\newline
    \textbf{Transformer (Text2text):} \citet{hoang2019efficient} is a summarization system using a pretrained Transformer.\newline
    \textbf{GraphWriter (Data2text):} \citet{koncel-kedziorski-etal-2019-text} is a graph transformer based model, which generates text using a seed title and a knowledge graph. Takes the database $X_\mathcal{G}$ as input. \newline
    \textbf{Entity (Data2text):} \citet{puduppully2019data} is an entity based data2text model, takes $X_\mathcal{G}$ as input.
\paragraph{Implementation Details}
 Our policy network is a three layer feedforward neural network.  We use a Transformer \citep{vaswani2017attention} implementation for Surface Realization. 
We train an off-the-shelf Neural CRF tagger \citep{yang2018ncrf++} for entity extraction. We use BERT \citep{devlin2018bert} based classifiers to predict the relation between two entities in a text trained using crowdsourced annotations from \citep{shah2021nutri}. Futher implementation details can be found in \ref{appendix:implementation_details}.

\section{Results}

\begin{table*}
\centering
\footnotesize
\scalebox{0.8}{
\begin{tabularx}{\textwidth}{c}
   \toprule
   Transformer (baseline) \\
   * Whole - grain cereals may protect against obesity , diabetes and certain cancers. However , more research is needed . \\
   * Whole grains , such as mozambican grass , are safe to eat with no serious side effects . \\
   * \color{red}{Whole - grain cereals may protect against obesity , diabetes and certain cancers. However , more research is needed .} \\
   * \color{red}{Whole grains , such as blueberries ,} \color{black}{are likely safe to eat with no serious side effects .}\\
   * Whole grains are safe to eat. \color{red}{However , people with type 2 diabetes should avoid whole grains .} \\
   * Whole grains are lower in carbs than whole grains , making them a good choice for people with type 2 diabetes. \\
    \midrule
\end{tabularx}}
\scalebox{0.8}{
\begin{tabularx}{\textwidth}{c}
  \textbf{Our Method} \\
  * Whole grains has been shown to lower weight gain and improve various type 2 diabetes risk factors .\\
  * Whole grains has been shown to lower insulin resistance and improve various cancer risk factors .\\
  * Whole grains has been linked to several other potential health benefits , such as improved CVD risk , eyesight , \\
  and memory. 
  However , more studies are needed to draw stronger conclusions.\\
  * There is some evidence , in both animals and humans , \\that whole grains can reduce mortality by regulating the
  hormone ghrelin.\\
  \bottomrule
\end{tabularx}
}
\caption{Example outputs of our model and the Transformer baseline for a multi-issues summary. Trained on limited parallel data, the Transformer baseline produces \color{red}{repetitive text} with \color{red}{factual inaccuracies}, \color{black}{while our method is able to provide more accurate and diverse summarization.}}
\label{tab:multicasestudy}
\end{table*}

In this section, we describe the performance of our \emph{Nutribullet Hybrid} system and baselines on  summarization and summary updates. We report empirical results , human evaluation and present sample outputs, highlighting the benefits of our method. 

\begin{table}[!htbp]
\small
\centering
\begin{tabular} {l|c|c} 
\toprule
Model        & KG(G) & KG(I)  \\
\midrule
Copy-gen               & 0.43 & 0.69\\
Transformer           & 0.33 & 0.73 \\
\midrule
Ours        & \textbf{0.5} & \textbf{0.90}\\
\bottomrule
\end{tabular}
\caption{KG in gold(G) and KG in input(I) in our model and baselines in the food and multi-issues setting . The best results are in \textbf{bold}.}
\label{tab:automatic-full2}
\end{table}

\textbf{Single and Multi-issues Summarization:} We describe the results on the task of generating summaries. Table ~\ref{tab:automatic-full} presents the automatic evaluation results for the food and single issue summarization task. High KG(I) and KG(G) scores for our method indicate that the generated text is faithful to input entities and relevant. In particular, a high Aggregation Cognisance (Ag) score indicates that our model generates summaries which are cognizant of the varying degrees of consensus in the input Pubmed documents. Compared to other baselines we also receive a competitive score on the automatic Rouge metric, beating Copy-gen, Entity Data2text and GraphWriter baselines while falling  short (by 1.7\%) of the Transformer baseline. The baselines, especially  Transformer, tend to produce similar outputs for different inputs (see Table \ref{tab:multicasestudy}). Since a lot of these patterns are learned from the human summaries, Transformer receives a high Rouge score. However, as in the low resource regime, the baseline does not completely capture the content and aggregation, it fails to get a very high KG(G) or Ag score. A similar trend is observed for the other baselines too, which in this low resource regime produce a lot of false information, reflected in their low KG(I) scores. 

Human evaluation, conducted by considering scores,on a 1-4 Likert scale, from three annotators for each instance, shows the same pattern. Our model is able to capture the most relevant information, when compared against the gold summaries while producing fluent summaries. The Transformer baseline produces fluent summaries, which are not as relevant. The performance is poorer for the Copy-gen, Entity Data2text and GraphWriter models.

In the multi-issues setting, the baselines access the gold annotations with respect to the input documents' clustering. Our model conducts the extra task of grouping the selected tuples, using the "New List" action. Our model performs better than the baselines on both the KG(I) and KG(G) metrics as seen in Table \ref{tab:automatic-full2}. Again, the pattern of producing very similar and repetitive sentences hurts the baselines. They fail to cover different issues and tend to produce false information, in this low resource setting. Our model scores an 7\% higher on KG(G) and 17\% higher on KG(I) compared to the next best performance, in absolute terms. Table \ref{tab:multicasestudy} shows the comparison between the outputs produced by our method and the Transformer baseline on the benefits of whole-grains. Our method conveys more relevant, factual and organized information in a concise manner.

\begin{table}[!htbp]
\small
\centering
\scalebox{0.85}{
\begin{tabular} {l|c|c} 
\toprule
\multicolumn{1}{c}{}&\multicolumn{1}{c}{Fusion Update} & \multicolumn{1}{c}{Contradictory Update}\\
Model   & KG(G) & Ag \\
\midrule
Copy-gen             &  0.16 & 0.50\\
GraphWriter          & 0.0  & 0.50\\
Entity Data2text & 0.16 & 0.50 \\
Transformer          &  0.16 & 0.46 \\
\midrule
Ours        &  \textbf{0.33} & \textbf{0.76} \\
\bottomrule
\end{tabular}
}
\caption{ The middle column shows KG in gold(G) in our model and baselines for fusion updates . The last column shows Aggregation Cognisance (Ag) in our model and baselines in the contradictory update setting. The best results are in \textbf{bold}.}
\label{tab:update_fusion}
\end{table}

\begin{table*}[!htbp]
\centering
\footnotesize
\scalebox{0.8}{
\begin{tabularx}{\textwidth}{p{1.8cm}Xl}
    \toprule
    Old Summary     & Flax seeds contain a group of nutrients called lignans , which have powerful antioxidant and estrogen properties . \\
     New Inputs  & (i):"...current overall evidence indicates that FS and its components are effective in the \color{blue}{risk reduction and treatment of breast cancer} \color{black}{and safe for consumption by breast cancer patients..." (ii): "...Consumption of flaxseed was associated with a} \color{blue}{significant reduction in breast cancer risk} \color{black}{as was consumption of flax bread ..." (iii): "...a flaxseed-supplemented, fat-restricted diet may affect the biology of the prostate and associated biomarkers..."} \\
    \midrule
\end{tabularx}
}
\scalebox{0.8}{
\begin{tabularx}{\textwidth}{p{1.8cm}Xl}
    Copy-gen & Avocados may help fight cancer risk, boost inflammation. In a pasteurized called polyphenols, which may aid weight loss.\\
    Transformer & Flaxseed oil is high in antioxidants that may help reduce the risk of several chronic diseases . \\
    Ours  &  Flax seeds are rich in antioxidant  , especially through lignans. They contain beneficial nutrients which can \color{blue}{help protect your body against certain types of breast cancer} .
\\    \midrule
\end{tabularx}
}
\scalebox{0.8}{
\begin{tabularx}{\textwidth}{p{1.8cm}Xl}
    \midrule
    Old Summary     & Flax seeds, high in fiber, can be a beneficial addition to the diet of people with diabetes . \\
     New Input  & "...showed fasting \color{blue}{blood sugar in the experimental group decreased}\color{black}{...the total cholesterol reduced...Results showed} \color{blue}{a decrease in low-density lipoprotein cholesterol}\color{black}{...The study demonstrated the efficacy of flax gum in the blood biochemistry profiles of type 2 diabetes."} \\
    \midrule
\end{tabularx}
}
\scalebox{0.8}{
\begin{tabularx}{\textwidth}{p{1.8cm}Xl}
    Copy-gen & Eating apart has been linked to increased growth cholesterol, and cholesterol levels. However, more studies are needed to confirm possible effect.\\
    Transformer & \color{red}{Flaxseed extract may}\color{blue}{ help lower blood sugar levels .} \\
    Ours  &  Flax seeds are high in fiber , which is beneficial for people with diabetes and associated with a \color{blue}{reduced  low-density lipoprotein cholesterol} .
\\    \bottomrule
\end{tabularx}
}
\caption{Example outputs of our model and baselines for a summary update upon receiving \color{blue}{new information} \color{black}{about flaxseeds + cancer and flaxseeds + cholesterol, respectively.} Our model maintains old information and updates accurately. In the cholesterol case, Transformer adds new information \color{red}{but misses the old information.}}
\label{tab:fusioncasestudy}
\end{table*}

\textbf{Summary Update:}
We study the efficacy of our model to fuse information in existing summaries on receiving new Pubmed studies. As the KG(G) metric in 
\ref{tab:update_fusion} shows, our model is able to select and fuse more relevant information. Table \ref{tab:fusioncasestudy} shows two examples of summaries on flaxseeds where our model successfully fuses new information. 

Table \ref{tab:update_fusion}'s last column presents the automatic evaluation results to demonstrate the efficacy of maintaining Aggregation Cognisance (Ag), which is critical when updating summaries on receiving contradictory results. The high performance in this update setting demonstrates the \emph{Surface Realization} model's ability to produce aggregation cognizant outputs, in contrast to the baselines that do not learn this reasoning in a low resource regime.


\textbf{Analysis: Information Extraction and Content Aggregation} Information extraction is the critical first step performed for the input documents in order to get symbolic data for content selection and aggregation. To this end, we report the performance of the information extraction system, which is composed of two models -- entity extraction and relation classification. As reported in Table \ref{tab:ie_aggregation}, the entity extraction model, a crf-based sequence tagging model, receives a token-level F1 score of 79\%. The relation classification model, a BERT based text classifier, receives an accuracy of 69\%.

The performance of the information extraction models is particularly important for the content aggregation sub-task. In order to analyse this quantitatively, we perform manual analysis of the 179 instances in the dev set and compare them to the system identified aggregation -- information extraction followed by the deterministic rules in Table \ref{table:rules}. Given the simplicity of our rules, system's 78\% accuracy in Table \ref{tab:ie_aggregation} is acceptable. Deeper analysis shows that the performance is lowest for Population Scoping and Contradiction with an accuracy of 52\% and 56\% respectively. The performance of Population Scoping being low is down predominantly to the simplicity of the rules. Most mistakes occur when the input studies are review studies that don't mention any population but analyze results from several past work. Contradiction suffers because of the information extraction system and stronger models for the same should be able to alleviate the errors.

\begin{table}[!htbp]
\small
\centering
\scalebox{0.85}{
\begin{tabular} {l|c} 
\toprule
Task   & Performance  \\
\midrule
Entity Extraction             &  0.79\\
Relation Classification &  0.69 \\
Aggregation Operator Identification        &  0.78 \\
\bottomrule
\end{tabular}
}
\caption{ Performance of our information extraction system and its impact on content aggregation.}
\label{tab:ie_aggregation}
\end{table}

\section{Conclusion}
 While modern models produce fluent text in multi-document summarization, they struggle to capture the consensus amongst the input documents. This inadequacy -- magnified in low resource domains, is addressed by our model. Our model is able to generate robust summaries which are faithful to content and cognizant of the varying consensus in the input documents. Our approach is applicable in summarization and textual updates.  Extensive experiments, automatic and human evaluation underline its impact over state-of-the-art baselines.
\section*{Acknowledgements}
We thank the MIT NLP group and the reviewers for their helpful discussion and comments. This work is supported by DSO grant DSOCO1905.

\bibliography{mybib}

\begin{thebibliography}{35}
\expandafter\ifx\csname natexlab\endcsname\relax\def\natexlab#1{#1}\fi

\bibitem[{Albaum(1997)}]{albaum1997likert}
Gerald Albaum. 1997.
\newblock The likert scale revisited.
\newblock \emph{Market Research Society. Journal.}, 39(2):1--21.

\bibitem[{Amplayo and Lapata(2019)}]{amplayo2019informative}
Reinald~Kim Amplayo and Mirella Lapata. 2019.
\newblock \href {https://arxiv.org/abs/1911.02247} {Informative and
  controllable opinion summarization}.
\newblock \emph{arXiv preprint arXiv:1909.02322}.

\bibitem[{Barzilay et~al.(1998)Barzilay, McCullough, Rambow, DeCristofaro,
  Korelsky, and Lavoie}]{barzilay-etal-1998-new}
Regina Barzilay, Daryl McCullough, Owen Rambow, Jonathan DeCristofaro, Tanya
  Korelsky, and Benoit Lavoie. 1998.
\newblock \href {https://www.aclweb.org/anthology/W98-1409} {A new approach to
  expert system explanations}.
\newblock In \emph{Natural Language Generation}.

\bibitem[{Baumel et~al.(2018)Baumel, Eyal, and Elhadad}]{baumel2018query}
Tal Baumel, Matan Eyal, and Michael Elhadad. 2018.
\newblock \href {https://arxiv.org/abs/1801.07704} {Query focused abstractive
  summarization: Incorporating query relevance, multi-document coverage, and
  summary length constraints into seq2seq models}.
\newblock \emph{arXiv preprint arXiv:1801.07704}.

\bibitem[{Cheng and Lapata(2016)}]{cheng-lapata-2016-neural}
Jianpeng Cheng and Mirella Lapata. 2016.
\newblock \href {https://doi.org/10.18653/v1/P16-1046} {Neural summarization by
  extracting sentences and words}.
\newblock In \emph{Proceedings of the 54th Annual Meeting of the Association
  for Computational Linguistics (Volume 1: Long Papers)}, pages 484--494,
  Berlin, Germany. Association for Computational Linguistics.

\bibitem[{Devlin et~al.(2018)Devlin, Chang, Lee, and
  Toutanova}]{devlin2018bert}
Jacob Devlin, Ming-Wei Chang, Kenton Lee, and Kristina Toutanova. 2018.
\newblock \href {https://arxiv.org/abs/1810.04805} {Bert: Pre-training of deep
  bidirectional transformers for language understanding}.
\newblock \emph{arXiv preprint arXiv:1810.04805}.

\bibitem[{Donahue et~al.(2020)Donahue, Lee, and Liang}]{donahue2020enabling}
Chris Donahue, Mina Lee, and Percy Liang. 2020.
\newblock Enabling language models to fill in the blanks.
\newblock \emph{arXiv preprint arXiv:2005.05339}.

\bibitem[{Fabbri et~al.(2019)Fabbri, Li, She, Li, and Radev}]{multinews}
Alexander Fabbri, Irene Li, Tianwei She, Suyi Li, and Dragomir Radev. 2019.
\newblock \href {https://doi.org/10.18653/v1/P19-1102} {Multi-news: A
  large-scale multi-document summarization dataset and abstractive hierarchical
  model}.
\newblock In \emph{Proceedings of the 57th Annual Meeting of the Association
  for Computational Linguistics}, pages 1074--1084, Florence, Italy.
  Association for Computational Linguistics.

\bibitem[{Fan et~al.(2017)Fan, Grangier, and Auli}]{fan2017controllable}
Angela Fan, David Grangier, and Michael Auli. 2017.
\newblock Controllable abstractive summarization.
\newblock \emph{arXiv preprint arXiv:1711.05217}.

\bibitem[{Guu et~al.(2018)Guu, Hashimoto, Oren, and
  Liang}]{guu-etal-2018-generating}
Kelvin Guu, Tatsunori~B. Hashimoto, Yonatan Oren, and Percy Liang. 2018.
\newblock \href {https://doi.org/10.1162/tacl_a_00030} {Generating sentences by
  editing prototypes}.
\newblock \emph{Transactions of the Association for Computational Linguistics},
  6:437--450.

\bibitem[{Guu et~al.(2017)Guu, Pasupat, Liu, and
  Liang}]{guu-etal-2017-language}
Kelvin Guu, Panupong Pasupat, Evan Liu, and Percy Liang. 2017.
\newblock \href {https://doi.org/10.18653/v1/P17-1097} {From language to
  programs: Bridging reinforcement learning and maximum marginal likelihood}.
\newblock In \emph{Proceedings of the 55th Annual Meeting of the Association
  for Computational Linguistics (Volume 1: Long Papers)}, pages 1051--1062,
  Vancouver, Canada. Association for Computational Linguistics.

\bibitem[{Hoang et~al.(2019)Hoang, Bosselut, Celikyilmaz, and
  Choi}]{hoang2019efficient}
Andrew Hoang, Antoine Bosselut, Asli Celikyilmaz, and Yejin Choi. 2019.
\newblock \href {https://arxiv.org/abs/1906.00138} {Efficient adaptation of
  pretrained transformers for abstractive summarization}.
\newblock \emph{arXiv preprint arXiv:1906.00138}.

\bibitem[{Koncel-Kedziorski et~al.(2019)Koncel-Kedziorski, Bekal, Luan, Lapata,
  and Hajishirzi}]{koncel-kedziorski-etal-2019-text}
Rik Koncel-Kedziorski, Dhanush Bekal, Yi~Luan, Mirella Lapata, and Hannaneh
  Hajishirzi. 2019.
\newblock \href {https://doi.org/10.18653/v1/N19-1238} {{T}ext {G}eneration
  from {K}nowledge {G}raphs with {G}raph {T}ransformers}.
\newblock In \emph{Proceedings of the 2019 Conference of the North {A}merican
  Chapter of the Association for Computational Linguistics: Human Language
  Technologies, Volume 1 (Long and Short Papers)}, pages 2284--2293,
  Minneapolis, Minnesota. Association for Computational Linguistics.

\bibitem[{Kry{\'s}ci{\'n}ski et~al.(2019)Kry{\'s}ci{\'n}ski, McCann, Xiong, and
  Socher}]{kryscinski2019evaluating}
Wojciech Kry{\'s}ci{\'n}ski, Bryan McCann, Caiming Xiong, and Richard Socher.
  2019.
\newblock \href
  {https://ui.adsabs.harvard.edu/abs/2019arXiv191012840K/abstract} {Evaluating
  the factual consistency of abstractive text summarization}.
\newblock \emph{arXiv}, pages arXiv--1910.

\bibitem[{Lebanoff et~al.(2018)Lebanoff, Song, and Liu}]{lebanoff2018adapting}
Logan Lebanoff, Kaiqiang Song, and Fei Liu. 2018.
\newblock \href {https://arxiv.org/abs/1808.06218} {Adapting the neural
  encoder-decoder framework from single to multi-document summarization}.
\newblock \emph{arXiv preprint arXiv:1808.06218}.

\bibitem[{Lewis et~al.(2019)Lewis, Liu, Goyal, Ghazvininejad, Mohamed, Levy,
  Stoyanov, and Zettlemoyer}]{lewis2019bart}
Mike Lewis, Yinhan Liu, Naman Goyal, Marjan Ghazvininejad, Abdelrahman Mohamed,
  Omer Levy, Ves Stoyanov, and Luke Zettlemoyer. 2019.
\newblock \href {https://arxiv.org/abs/1910.13461} {Bart: Denoising
  sequence-to-sequence pre-training for natural language generation,
  translation, and comprehension}.
\newblock \emph{arXiv preprint arXiv:1910.13461}.

\bibitem[{Li(2018)}]{li-2018-learning}
Yanpeng Li. 2018.
\newblock \href {https://www.aclweb.org/anthology/C18-1241} {Learning features
  from co-occurrences: A theoretical analysis}.
\newblock In \emph{Proceedings of the 27th International Conference on
  Computational Linguistics}, pages 2846--2854, Santa Fe, New Mexico, USA.
  Association for Computational Linguistics.

\bibitem[{Lin(2004)}]{lin2004rouge}
Chin-Yew Lin. 2004.
\newblock Rouge: A package for automatic evaluation of summaries.
\newblock In \emph{Text summarization branches out}, pages 74--81.

\bibitem[{McKeown and Radev(1995)}]{mckeown1995generating}
Kathleen McKeown and Dragomir~R Radev. 1995.
\newblock \href {https://dl.acm.org/doi/pdf/10.1145/215206.215334} {Generating
  summaries of multiple news articles}.
\newblock In \emph{Proceedings of the 18th annual international ACM SIGIR
  conference on Research and development in information retrieval}, pages
  74--82.

\bibitem[{Mueller et~al.(2017)Mueller, Gifford, and
  Jaakkola}]{mueller2017sequence}
Jonas Mueller, David Gifford, and Tommi Jaakkola. 2017.
\newblock Sequence to better sequence: continuous revision of combinatorial
  structures.
\newblock In \emph{International Conference on Machine Learning}, pages
  2536--2544.

\bibitem[{Puduppully et~al.(2019)Puduppully, Dong, and
  Lapata}]{puduppully2019data}
Ratish Puduppully, Li~Dong, and Mirella Lapata. 2019.
\newblock \href {https://arxiv.org/abs/1906.03221} {Data-to-text generation
  with entity modeling}.
\newblock \emph{arXiv preprint arXiv:1906.03221}.

\bibitem[{Radev and McKeown(1998)}]{radev-mckeown-1998-generating}
Dragomir~R. Radev and Kathleen~R. McKeown. 1998.
\newblock \href {https://www.aclweb.org/anthology/J98-3005} {Generating natural
  language summaries from multiple on-line sources}.
\newblock \emph{Computational Linguistics}, 24(3):469--500.

\bibitem[{Rush et~al.(2015)Rush, Chopra, and Weston}]{rush-etal-2015-neural}
Alexander~M. Rush, Sumit Chopra, and Jason Weston. 2015.
\newblock \href {https://doi.org/10.18653/v1/D15-1044} {A neural attention
  model for abstractive sentence summarization}.
\newblock In \emph{Proceedings of the 2015 Conference on Empirical Methods in
  Natural Language Processing}, pages 379--389, Lisbon, Portugal. Association
  for Computational Linguistics.

\bibitem[{See et~al.(2017)See, Liu, and Manning}]{see-etal-2017-get}
Abigail See, Peter~J. Liu, and Christopher~D. Manning. 2017.
\newblock \href {https://doi.org/10.18653/v1/P17-1099} {Get to the point:
  Summarization with pointer-generator networks}.
\newblock In \emph{Proceedings of the 55th Annual Meeting of the Association
  for Computational Linguistics (Volume 1: Long Papers)}, pages 1073--1083,
  Vancouver, Canada. Association for Computational Linguistics.

\bibitem[{Shah et~al.(2019)Shah, Schuster, and Barzilay}]{shah2019automatic}
Darsh~J Shah, Tal Schuster, and Regina Barzilay. 2019.
\newblock \href {https://arxiv.org/abs/1909.13838} {Automatic fact-guided
  sentence modification}.
\newblock \emph{arXiv preprint arXiv:1909.13838}.

\bibitem[{Shah et~al.(2021)Shah, Yu, Lei, and Barzilay}]{shah2021nutri}
Darsh~J Shah, Lili Yu, Tao Lei, and Regina Barzilay. 2021.
\newblock \href {https://arxiv.org/abs/2103.11921} {Nutri-bullets: Summarizing
  health studies by composing segments}.
\newblock \emph{arXiv preprint arXiv:2103.11921}.

\bibitem[{Sharma et~al.(2019)Sharma, Huang, Hu, and
  Wang}]{sharma-etal-2019-entity}
Eva Sharma, Luyang Huang, Zhe Hu, and Lu~Wang. 2019.
\newblock \href {https://doi.org/10.18653/v1/D19-1323} {An entity-driven
  framework for abstractive summarization}.
\newblock In \emph{Proceedings of the 2019 Conference on Empirical Methods in
  Natural Language Processing and the 9th International Joint Conference on
  Natural Language Processing (EMNLP-IJCNLP)}, pages 3280--3291, Hong Kong,
  China. Association for Computational Linguistics.

\bibitem[{Shen et~al.(2020)Shen, Quach, Barzilay, and Jaakkola}]{shen2020blank}
Tianxiao Shen, Victor Quach, Regina Barzilay, and Tommi Jaakkola. 2020.
\newblock \href {https://arxiv.org/abs/2002.03079} {Blank language models}.
\newblock \emph{arXiv preprint arXiv:2002.03079}.

\bibitem[{Stern et~al.(2019)Stern, Chan, Kiros, and
  Uszkoreit}]{stern2019insertion}
Mitchell Stern, William Chan, Jamie Kiros, and Jakob Uszkoreit. 2019.
\newblock \href {https://arxiv.org/abs/1902.03249} {Insertion transformer:
  Flexible sequence generation via insertion operations}.
\newblock \emph{arXiv preprint arXiv:1902.03249}.

\bibitem[{Vaswani et~al.(2017)Vaswani, Shazeer, Parmar, Uszkoreit, Jones,
  Gomez, Kaiser, and Polosukhin}]{vaswani2017attention}
Ashish Vaswani, Noam Shazeer, Niki Parmar, Jakob Uszkoreit, Llion Jones,
  Aidan~N Gomez, {\L}ukasz Kaiser, and Illia Polosukhin. 2017.
\newblock Attention is all you need.
\newblock In \emph{Advances in neural information processing systems}, pages
  5998--6008.

\bibitem[{Wenbo et~al.(2019)Wenbo, Yang, Heyan, and Yuxiang}]{wenbo2019concept}
Wang Wenbo, Gao Yang, Huang Heyan, and Zhou Yuxiang. 2019.
\newblock \href {https://arxiv.org/abs/1910.08486} {Concept pointer network for
  abstractive summarization}.
\newblock \emph{arXiv preprint arXiv:1910.08486}.

\bibitem[{Williams(1992)}]{williams1992simple}
Ronald~J Williams. 1992.
\newblock Simple statistical gradient-following algorithms for connectionist
  reinforcement learning.
\newblock \emph{Machine learning}, 8(3-4):229--256.

\bibitem[{Wiseman et~al.(2018)Wiseman, Shieber, and
  Rush}]{wiseman-etal-2018-learning}
Sam Wiseman, Stuart Shieber, and Alexander Rush. 2018.
\newblock \href {https://doi.org/10.18653/v1/D18-1356} {Learning neural
  templates for text generation}.
\newblock In \emph{Proceedings of the 2018 Conference on Empirical Methods in
  Natural Language Processing}, pages 3174--3187, Brussels, Belgium.
  Association for Computational Linguistics.

\bibitem[{Yang and Zhang(2018)}]{yang2018ncrf++}
Jie Yang and Yue Zhang. 2018.
\newblock \href {https://arxiv.org/abs/1806.05626} {Ncrf++: An open-source
  neural sequence labeling toolkit}.
\newblock \emph{arXiv preprint arXiv:1806.05626}.

\bibitem[{Zhang et~al.(2018)Zhang, Tan, and Wan}]{zhang-etal-2018-adapting}
Jianmin Zhang, Jiwei Tan, and Xiaojun Wan. 2018.
\newblock \href {https://doi.org/10.18653/v1/W18-6545} {Adapting neural
  single-document summarization model for abstractive multi-document
  summarization: A pilot study}.
\newblock In \emph{Proceedings of the 11th International Conference on Natural
  Language Generation}, pages 381--390, Tilburg University, The Netherlands.
  Association for Computational Linguistics.

\end{thebibliography}
\bibliographystyle{acl_natbib}

\clearpage
\appendix
\section{Appendix}
\label{sec:appendix}
\renewcommand\thefigure{\thesection.\arabic{figure}}    
\setcounter{figure}{0}

\begin{table}
\center
\scalebox{1.0}{
\begin{tabular}{l|c}
\toprule
Aggregation Operator & Deterministic Rule\\
\midrule
Population Scoping &   $|$Specific Population$|$ < Threshold\\
Contradiction &  "evidence is mixed", "conflicting" or "contradiction" in $y_m$  \\
Under-Reported &  "more research is needed", "more studies are needed" or "more human studies" in $y_m$  \\
 Agreement & None of the Above \\
\bottomrule
\end{tabular}
}
\caption{Deterministic Rules to identify the Aggregation Operator on outputs.}
\label{table:rules_output}
\end{table}

\paragraph{Implementation Details:}
\label{appendix:implementation_details}
The hyper-parameters for the content selection model are shared along with the code. The hyper-parameters for surface realization \citep{lewis2019bart} as the default values present in \url{https://github.com/pytorch/fairseq/blob/master/examples/bart/README.md}.

\paragraph{Baselines}
We use publicly available implementations for all our baselines. Copy-gen is from \url{https://github.com/atulkum/pointer_summarizer}. GraphWriter is from \url{https://github.com/rikdz/GraphWriter}. The Transformer for abstractive summarization, pretrained implementation is from \url{https://github.com/Andrew03/transformer-abstractive-summarization}. Entity Data2text implementation is the closest working and usable implementation of the model in Opennmt-py \url{https://github.com/OpenNMT/OpenNMT-py/blob/master/config/config-rnn-summarization.yml}. We provide the full database $X_\mathcal{G}$ as input along with the food name.

\paragraph{Aggregation Operator for Outputs}

We specify our rule based classifier for summary $y_m$'s aggregation operator $O^y_m$ identification. The following patterns are checked.

\label{sec:supplemental}

\end{document}